# STRATEGIC LEARNING AND ROBUST PROTOCOL DESIGN FOR ONLINE COMMUNITIES WITH SELFISH USERS

Yu Zhang, Mihaela van der Schaar
Department of Electrical Engineering, UCLA
yuzhang@ucla.edu, mihaela@ee.ucla.edu

***Abstract*****—***This paper focuses on analyzing the free-riding behavior of self-interested users in online communities. Hence, traditional optimization methods for communities composed of compliant users such as network utility maximization cannot be applied here. In our prior work, we show how social reciprocation protocols can be designed in online communities which have populations consisting of a continuum of users and are stationary under stochastic permutations. Under these assumptions, we are able to prove that users voluntarily comply with the pre-determined social norms and cooperate with other users in the community by providing their services. In this paper, we generalize the study by analyzing the interactions of self-interested users in online communities with finite populations and are not stationary. To optimize their long-term performance based on their knowledge, users adapt their strategies to play their best response by solving individual stochastic control problems. The best-response dynamic introduces a stochastic dynamic process in the community, in which the strategies of users evolve over time. We then investigate the long-term evolution of a community, and prove that the community will converge to stochastically stable equilibria which are stable against stochastic permutations. Understanding the evolution of a community provides protocol designers with guidelines for designing social norms in which no user has incentives to adapt its strategy and deviate from the prescribed protocol, thereby ensuring that the adopted protocol will enable the community to achieve the optimal social welfare.*

## I. INTRODUCTION

The proliferation of social networking services has permeated our social and economic lives and created online social communities where individuals interact with each other. Examples of online social communities include peer-to-peer applications [1][2][3], online crowd-sourcing services [4], grid computing [5], as well as social media services [6]. Such communities play an increasingly central role in enabling individual users to remotely share and obtain job opportunities, trade goods and services, acquire knowledge and learn new trends etc. However, online communities in general rely on voluntary contribution of services by individual users, and are therefore vulnerable to intrinsic incentive problems which lead to prevalent free-riding behaviors among users, at the expense of the collective social welfare of the community [1][2].

Various incentive mechanisms have been proposed to encourage cooperation in online communities [2][3][6], with a large body of them relying on the idea of *reciprocity*, in which users are rated based on their past behaviors. Differential service schemes which provide services based on users' rating scores are deployed to encourage users to contribute to the community in order to receive better services in return. Reciprocity-based mechanisms can be further classified into *direct reciprocity* and *indirect reciprocity* depending on how the rating score is generated. Existing direct reciprocity protocols [2][3], in which a user is rated independently by each other user whom it has interactions with, do not scale well to online communities with large populations of anonymous users that are randomly matched [1][4]. This is due to the fact that frequent interactions between two users, which are required in order to build up accurate mutual ratings in this bilateral reciprocation paradigm, are absent. As a result, in online communities with large populations of users, a new paradigm based on indirect reciprocity [8], in which a user is rated collectively by all other users it has interactions with, is necessary to compel users not to free-ride [9][10].

In our prior work [11], we have developed an indirect reciprocity framework based on social norms in order to incentivize self-interested users to cooperate in online communities. Social norms are defined as the rules that are deployed by a protocol designer to regulate the users' behaviors. A label is assigned to each user indicating its reputation, which contains information about its past behavior. Users with different labels are treated differently by other users with whom they interact and the differential treatment is prescribed by a social rule designed in the social norm. Hence, an individual user can be rewarded or punished by other individuals in a community who have not had direct interactions with it. Since indirect reciprocation within the framework of social norms requires neither observable identities nor frequent interactions, it can be easily adopted in various online communities as long as an infrastructure, such as a tracker or a portal, exists for collecting, processing, and delivering information about the users' behavior.

In the analysis of [11], the state of a community is represented by the reputation distribution of its users. A model with the population consisting of a continuum of users is considered to characterize a community where the number of users is large. According to the law of large numbers [12], the reputation distribution of the community remains at a stationary point and thus, its long term average is stable under stochastic permutations, e.g. operation errors, noises, etc., since the changes in individual users' reputations average out at the community level in the continuum population. Hence, the interactions among users in [11] can be formulated as a mean-field game [13], and the social norm equilibrium only enforces users to comply with the protocol in the scenario where each user optimizes its individual utility only with respect to the stationary state.

Nevertheless, in a real community with a finite population where the law of large numbers does not hold, the state of the community is subject to stochastic permutations and no longer remains stable but varies over time and hence, the mean-field theory [13] can no longer be applied. Such variation in the community's state will thus affect the users' incentives to comply with the protocol designed in [11], which only enforces users' compliance in a particular state. For example, in P2P networks, a peer might provide upload services voluntarily if there are many peers who reciprocate by providing services in return; whereas, in a network where free-riders are the majority, a peer will choose to change its sharing behavior and adopt a more selfish strategy. Therefore, it is of critical importance to understand how a community evolves and transits between different states and where it finally converges to starting from an arbitrary state under a social norm, which were neglected in [11]. Such an analysis can provide essential insights that can inform the design of more efficient and robust incentive protocols for online communities.

A community evolves with users learning the environment and adapting their strategies over time. Such adaptations have been modeled as multi-agent learning (MAL) problems, which have been studied in various areas: artificial intelligence, evolutionary biology, control theory, and etc. The MAL research can be mainly categorized into two agendas: descriptive learning and prescriptive learning [14].

In descriptive learning [15][16][17], users do not prescribe any protocol or artificial learning rules. The goal of these works is to investigate formal models of learning that agree with people's natural behavior. In general, they ask the question that how natural users learn in the context of other learners and thus compute properties of the resulting game, e.g. equilibrium results and convergence of the learning and adaptation process. For example, the Bellman-style single-agent reinforcement learning techniques (in particular, Q-learning) is employed in the multi-agent setting [15][16]. The authors analyze whether the strategy of an individual user who adopts Q-learning converges to a stationary point and, if it converges, what is the structure of the strategy. The works on descriptive learning, however, fail to answer the question on how the protocol designers should design protocols in order to construct a multi-agent system to achieve some desirable properties, e.g. the highest social welfare in an online community.

This question, on the other hand, is answered in the works on prescriptive learning, which exploit how users should learn and how the protocol designers should construct protocols that induct users to learn certain strategies to achieve certain goals [18][19][20][21]. However, the current research in prescriptive learning mainly focuses on the cooperative games where either the interests of the protocol designers and the interests of the learning users are aligned [19][20][21], or users are exogenously compliant and have no freedom to deviate from the prescribed learning rules that often contradict their self-interests [18].

Our paper possesses a different goal than the MAL research. Unlike [19][20][21], the interests of the protocol designer and of the self-interested users in our work are in conflict of each other. Therefore, the protocol designer has to consider the trade-off between maximizing the social welfare and providing sufficient incentives for users to comply with the protocol. Instead of restricting the users' learning and adaptation rules and assuming their obedience as in [18], we allow users to form their own beliefs depending on the observations they make based on their past interactions with the community and, based on these beliefs, adapt their service strategies using the best response to maximize their individual long-term utilities. By considering the users' best response dynamics, we focus on the design of effective protocols based on social norms which induce users learning in the long term to cooperate with each other and contribute their services to the community.

The adaptation process of a self-interested user is formulated as a Markov Decision Process (MDP) [22], and we first prove that a user's best response adaptation is always harmful to the community because it decreases the social welfare. Next, we study the evolution of the users' strategies, and how the limiting behaviors of users are affected by the design of social norms in the long term when time goes to infinity. To characterize the evolutions in the long term, we introduce the concept of *stochastically stable equilibrium* (SSE) [23]. Generally speaking, an SSE is a state of the community together with a strategy profile of users which is the best response to this state. The state is stationary under the best response dynamics and emerges in the long term with a positive probability in the community which is subject to stochastic permutations. Hence, when the community operates at an SSE, the state of the community remains stationary under the best-response dynamics and users will thus play fixed service strategies that do not change over time. Understanding how a community evolves over time thus enables us to design protocols based on social norms which induce the cooperation among self-interest users with the social welfare of the community being maximized. In addition, our work also studies how the design of protocols needs to be adjusted to different exogenous community characteristics, including the service cost and benefit, the population size of the community, and the users' discount factor for the future utility. In the final part, we extend our work to consider additional issues such as the long-term behavior of the community when users are heterogeneous and have different beliefs.

The remainder of this paper is organized as follows. In Section II, we introduce our proposed social norm based framework for indirect reciprocity in online communities and define the design problem which needs to be solved by the protocol designers. In Section III, we study the users' learning and adaptation behavior given the designed protocol and how they impact the long-term evolution of the community. With this knowledge, we then solve the design problem to find the optimal protocol that maximizes the social welfare. Section IV presents our experimental results and the conclusions are drawn in Section V where directions for future research are also outlined.

# II. System Model

*A. Network assumption*

We consider an online community consisting of $N$ users. Each user possesses (or can produce) some valuable resources which enable them to offer services to others, such as knowledge [7], data files [3][6], computation resources [5], storage space [24], etc. The community is modeled as a discrete-time system with time divided into periods. In each period, each user generates one service request and selects an idle user who is not serving others to request the service [2] [1]. For the simplicity of analysis, we assume that the resources are uniformly distributed among all users, and hence the selection can be approximated by a uniformly random matching process such that all users in the community have an equal probability to be chosen by a particular user. As a result of the random matching process, each user also receives one service request from other users per period [2].

In order to characterize the asymmetry of interests among users, we model the interaction between a pair of matched users as a one-period asymmetric gift-giving game [25]. The user who requests services is called a *client* and the user who is being requested is called a *server* [3]. Upon receiving the request, the server selects its level of contribution to the client. To make the analysis tractable, we assume that the server's choice on its contribution level is binary [4], which can be quantified by $z \in \mathcal{Z} = \{0,1\}$, where $z = 1$ indicates that the server provides the requested service to the client and $z = 0$ indicates that the server refuses to provide the service.

[1] Here we assume that if two requesting users select the same idle user, only one of them is accepted for services and the rejected user redirects its service request to another idle user.

[2] The analysis can be extended to a general case when the service request rate per period is a positive real number [31]. It is assumed to be 1 in this manuscript only for illustration purposes.

[3] It is worth being noted that each user is engaged in two transactions in each period, one as a server (being requested by others) and the other as a client (requesting services from others).

[4] The framework in this manuscript is also applicable to the case when the contribution level of servers is not binary.

The users' utilities in a one-period game are determined by the server's level of contribution. When the server provides a service by choosing $z = 1$, it incurs a cost of $c$, and the client gains a benefit of $b$ upon the reception of the requested service; whereas both users receive a utility of 0 when the server chooses $z = 0$. Here we assume $b > c$ such that the game is socially valuable. The utility matrix of a one-period gift-giving game is presented in Table 1, with the client's utility being the first element in each entry and the server's utility being the second element.

Table1. Utility matrix of a gift-giving game.

| | Server $z = 1$ | Server $z = 0$ |
|---|---|---|
| Client | $b, -c$ | $0, 0$ |

The social welfare of the community is quantified by the average one-period utility received by all users in the community, denoted as $U$. It is maximized at $U = b - c$ if all users are cooperative and choose $z = 1$ when being requested for services. Nevertheless, $z = 0$ is always chosen by a self-interested server if it expects to maximize its one-period utility myopically. Hence, the social welfare $U = 0$ at the Nash equilibrium of the one-period game with no user providing services to others. In order to improve the inefficiency of the myopic Nash equilibrium, the repeated nature of the users' interactions is exploited to provide incentives for cooperation by using the threat of future punishments. To formalize this idea, we design a new set of incentive protocols based on social norms. As discussed in Section I, a social norm proposes a set of rules to regulate users' behavior. Compliance to these rules is positively rewarded (i.e. an increased level of service is provided to such users in the future) and failure to follow these rules can result in (severe) punishments (i.e. a decreased level of service is provided). Therefore, by carefully designing the reward and punishment rules, social norms are suitable to establish incentives for self-interested users to be active in contributing their services in online communities.

*B. An illustrative example of P2P file-sharing services*

A Peer-to-Peer service for file sharing, e.g. Gnutella [1] and BitTorrent [3], is one of the typical examples for our network model described in Section II.A. The resources in a P2P network are files (e.g. media chunks) shared public by each peer and can be accessed by every other peer in the network. There are usually some third-party managing devices (e.g. trackers) existing in the P2P network who maintain and update periodically the file map that records the file possession of each peer and is responsible for helping a requester (i.e. a peer requesting certain files) find the list of peers who have the requested file. The requester will then randomly choose a peer from this list to send the file download request [3]. When two peers are matched, the server consumes a cost for uploading the requested file to the client (e.g. energy consumption), while the client receives certain benefits from downloading the file (e.g. video distortion reduction for receiving a video chunk in media-sharing P2P applications). If the shared files are homogeneous which have the similar sizes and bring the similar benefits, the model in Section II.A can be used as an approximation with $b$ and $c$ representing the average benefit and cost related to one piece of file, respectively. The heterogeneous scenario is also important and it forms a future topic of our research.

*C. Repeated game formulation*

In a community, each user plays the stage game described in Section II.A repeatedly with different opponents. The regulation of a social norm takes effect through its manipulation over the users' social status, as in [8]. In the repeated game, the social status of a user is represented with a reputation $\theta \in \Theta$. We consider the reputation set $\Theta$ to be nonempty and finite, which is denoted as $\Theta = \{0, 1, 2, \cdots, L\}$ where $L$ is the highest reputation that can be obtained by a user. The service strategy that a user adopts in the repeated game is reputation-based and represented as a mapping $\sigma : \Theta \to \mathcal{Z}$, where each term $\sigma(\tilde{\theta}) \in \mathcal{Z}$ is the contribution level of this user when it is matched with a client of reputation $\tilde{\theta} \in \Theta$. The set of strategies that can be chosen by a user is hence finite and denoted as $\Gamma$.

We consider a social norm $\kappa = (\phi, \tau)$ that consists of a social rule and a reputation scheme, with a schematic representation illustrated in Figure 1.

A social rule $\phi$ is a set of service strategies that specifies the approved behavior of users within the community. For each user, its reputation is increased when it complies with $\phi$ in its service strategy and is decreased otherwise. Alternatively, the social rule can be represented as a mapping $\phi : \Theta \times \Theta \to \mathcal{Z}$, with $\phi(\theta, \tilde{\theta}) = \sigma_\theta(\tilde{\theta})$ for all $\theta$ and $\tilde{\theta}$, where $\sigma_\theta$ is the service strategy that the social rule instructs a server of reputation $\theta$ to choose, in order to get rewarded. For illustration, we consider threshold-based service strategies in a social rule. That is, each strategy $\sigma_\theta$ can be characterized by a service threshold $h_\theta \in \{0, 1, \cdots, L+1\}$, $\forall \theta$. By adopting $\sigma_\theta$, a user of reputation $\theta$ provides services only to clients whose reputations are no less than $h_\theta$ [5]. Formally, a threshold-based strategy can be represented as follows:

$$\sigma_\theta(\tilde{\theta}) = \begin{cases} 1 & if\ \tilde{\theta} \geq h_\theta \\ 0 & if\ \tilde{\theta} < h_\theta \end{cases}. \tag{1}$$

As particular cases, the fully cooperative strategy with $h_\theta = 0$ instructs a user of reputation $\theta$ to provide services to all users unconditionally and is denoted as $\sigma_C$, whereas the fully non-cooperative strategy with $h_\theta = L + 1$ that provides no service to any of the users is denoted as $\sigma_D$.

As an example, we consider social rules which satisfy the following property in the design of protocols:

$$\phi(\theta, \tilde{\theta}) = \begin{cases} 1 & if\ \theta \geq h\ and\ \tilde{\theta} \geq h \\ 1 & if\ \theta < h \\ 0 & otherwise \end{cases}. \tag{2}$$

Hence the social rule instructs a user of reputation $\theta < h$ to provide services to all users in the community, i.e. $h_\theta = 0, \forall \theta < h$, while a user of reputation $\theta \geq h$ only have to provide services to users whose reputation is also no less than $h$, i.e. $h_\theta = h, \forall \theta \geq h$. In a community where all users follow the social rule, a user of reputation $\theta \geq h$ receives services from all other users, while a user of reputation $\theta < h$ only receives services from users whose reputations are also below $h$. Hence, differential services are provided to users of different reputations according to the value of $h$, which is called the *social threshold* for convenience. Users of reputations lower than the social threshold are regarded as "*bad users*", and users of reputation no less than the social threshold are regarded as "*good users*". It should be noted that when $h = 0$ or $h = L + 1$, $\sigma_\theta = \sigma_C$ for all $\theta$. Such social rule cannot be enforced among self-interested users as shown in [11], due to the fact that it cannot provide differential services to users of different reputations. Therefore, we restrict our attention in the design on social rules with the threshold $h \in \{1, \cdots, L\}$, without loss of generality.

[5]The existing social norm based protocols such as [26] are in fact special cases of the threshold-based strategy proposed here.

A reputation scheme $\tau$ updates a user's reputation based on its past behavior as a server. After a server takes an action (i.e. decides whether to contribute or not), its client reports its contribution level to some trustworthy third-party managing device in the community (e.g. the tracker in P2P networks), and the managing device updates the server's reputation according to $\tau$ and the client's report. Formally, a reputation scheme $\tau$ updates a server's reputation based on the reputations of the matched users and the reported contribution level of the server. It is represented as a mapping $\tau : \Theta \times \Theta \times \mathcal{Z} \to \Theta$, in which $\tau(\theta, \tilde{\theta}, z)$ is the reputation of the server in the next period given its current reputation $\theta$, the client's reputation $\tilde{\theta}$, and the server's reported contribution level $z \in \mathcal{Z}$. As an example, we consider the following simple reputation scheme in this paper:

$$\tau(\theta, \tilde{\theta}, z) = \begin{cases} \min\{L, \theta + 1\} & if\ z = \phi(\theta, \tilde{\theta}) \\ 0 & if\ z \neq \phi(\theta, \tilde{\theta}) \end{cases}. \tag{3}$$

In this scheme, the server's reputation is increased by 1 while not exceeding $L$ if the reported contribution level matches that specified by the social rule $\phi$. Otherwise, the server's reputation is set to $0$ any time it deviates from the social rule. In practice, a community is continually being subjected to small stochastic perturbations that arise due to various types of operation errors in the community. To formalize the effect of such perturbations on the evolution of the community, we assume that the client's report is subject to a small error probability $\varepsilon$ of being reversed. This error could be the transmission error of clients' reports back to the managing device, or the perception error on the client's opinion about the server's contribution. Particularly, $z = 0$ is received by the managing device with probability $\varepsilon$ while the server actually plays $z = 1$, and vice versa. It should be noted that the value of $\varepsilon$ is known to all users in the community using appropriate estimation and measurements.

*D. Stochastically stable equilibrium and the design problem*

In general, a user's expected utility in one period depends on its own strategy as well as other users' reputations and strategies in the community. In the beginning of each period, the managing device posts on a website or broadcasts to its users the reputation distribution of the community, which we refer to as the *community configuration* denoted as $\mu = \{n(\theta)\}_{\theta=0}^{L}$ where $n(\theta)$ represents the number of users of reputation $\theta$ in the community. Due to their limited processing capabilities and interactions with others, users can only form simple beliefs about the strategies deployed by other users [28]. It should be noted that each user receives one service request in each period and hence, its reputation will also be updated once per period. Therefore, we assume that each user maintains a belief that a user of reputation higher than 0 will comply with the social rule $\phi$ in the current period with the probability $1-\varepsilon$ and play $\sigma_D$ with the probability $\varepsilon$, since its reputation is newly increased due to its compliance to the social rule in the previous period. In contrast, a user of reputation 0 will play $\sigma_D$ with the probability $1-\varepsilon$ and comply with $\phi$ with the probability $\varepsilon$ since it has been newly punished due to its deviation in the previous period [28].

Given the belief, a user's expected one-period utility thus becomes a function which is jointly determined by its own strategy $\sigma \in \Gamma$, its own reputation $\theta$, the social norm $\kappa$, and the community configuration $\mu$. Since a user can never be matched with itself, it is more convenient to compute a user's expected one-period utility by employing the reputation distribution of all users other than itself, which we refer to as the *opponent configuration*. The opponent configuration for a user of reputation $\theta$ is denoted as $\eta_\theta = \left\{m_\theta\left(\theta'\right)\right\}_{\theta'=0}^{L}$, which preserve the relationship with the community configuration as $m_\theta\left(\theta'\right) = n\left(\theta'\right)$ for all $\theta' \neq \theta$ and $m_\theta\left(\theta\right) = n\left(\theta\right) - 1$. Although the community configuration is unique in each period, different users face different opponent configurations. To avoid confusion, we will use the terminology "configuration" to refer to "community configuration" by default in the rest of this paper.

Let $v_\kappa\left(\sigma, \theta, \mu\right)$ denote the expected one-period utility for a user of reputation $\theta$ playing the strategy $\sigma$ when the community configuration is $\mu$, we can compute it as:

$$
\begin{aligned}
&v_\kappa\left(\sigma,\theta,\mu\right) = \\
&\frac{1-\varepsilon}{N-1}\sum_{\theta'\neq 0} m_\theta\left(\theta'\right)\left[b\left(\theta',\theta,\phi\left(\theta',\theta\right)\right) - c\left(\theta,\theta',\sigma\left(\theta'\right)\right)\right] + \frac{1-\varepsilon}{N-1} m_\theta\left(0\right)\left[b\left(0,\theta,\sigma_D\left(\theta\right)\right) - c\left(\theta,0,\sigma\left(0\right)\right)\right] \\
&+\frac{\varepsilon}{N-1}\sum_{\theta'\neq 0} m_\theta\left(\theta'\right)\left[b\left(\theta',\theta,\sigma_D\left(\theta\right)\right) - c\left(\theta,\theta',\sigma\left(\theta'\right)\right)\right] + \frac{\varepsilon}{N-1} m_\theta\left(0\right)\left[b\left(0,\theta,\phi\left(0,\theta\right)\right) - c\left(\theta,0,\sigma\left(0\right)\right)\right]
\end{aligned}
\quad . \tag{4}
$$

Here $b\left(\theta',\theta,\phi\left(\theta',\theta\right)\right)$ is the one-period benefit which this user can receive when its matched server has a reputation $\theta'$ and complies with $\phi$; $b\left(\theta',\theta,\sigma_D\left(\theta\right)\right)$ is the one-period benefit which this user can receive when its matched server plays the strategy $\sigma_D$. $c\left(\theta,\theta',\sigma\left(\theta'\right)\right)$ is the one-period cost of this user when its matched client has a reputation $\theta'$. The expected one-period utility of a user complying with $\phi$ is compactly denoted as $v_\kappa\left(\theta,\mu\right) = v_\kappa\left(\sigma_\theta,\theta,\mu\right)$. A user's expected long-term utility in the repeated game is evaluated using the infinite-horizon discounted sum criterion. Starting from any period $t_0$, the user's expected long-term utility is expressed as

$$v_{\kappa}^{\infty}\left(\sigma^{(t_0)},\theta^{(t_0)},\mu^{(t_0)}\right)=E\left\{\sum_{t=t_0}^{\infty}\delta^{t-t_0}v_{\kappa}\left(\sigma^{(t)},\theta^{(t)},\mu^{(t)}\right)\right\}, \tag{5}$$

where $\delta \in [0,1)$ is the discount factor which represents a user's preference of the future utility.

Under a social norm $\kappa$, adaptive self-interested users play the best response in order to optimizes their expected long-term utilities. Hence, the best response $\sigma^*$ for a user in period $t_0$ is

$$\sigma^* = \arg\max_{\sigma} v_{\kappa}^{\infty}\left(\sigma,\theta^{(t_0)},\mu^{(t_0)}\right). \tag{6}$$

The best-response dynamic of users thus introduces a stochastic dynamic process in the community which will be analyzed in the next section using a Markov chain analysis. Users' strategies in this process evolve over time [23]. We are interested in whether this process may converge to an equilibrium in the long term, i.e. each user holds a fixed reputation and plays a fixed strategy after a sufficiently long period of time, and if an equilibrium exists, how the protocol designer can design the social norm in order to enforce all users to play cooperation in their best responses. This provides the protocol designer with guidelines for selecting the correct social norm based on the community characteristics, e.g. the utility structure $(b,c)$ and the discount factor $\delta$, in order to optimize the sharing efficiency in the community. To formalize the effect of stochastic permutations, which happen rarely in a community (i.e. with a small probability), we consider operation errors with the probability of occurrence approaches 0, i.e. $\varepsilon \to 0$. The resulting equilibrium is defined as a *stochastically stable equilibrium* (SSE) in [23]. Indexing all users in the community by $i \in \{1,\ldots,N\}$ and denote the reputation profile and the strategy profile of all users as $\boldsymbol{\theta} = \left(\theta^1,\ldots,\theta^i,\ldots,\theta^N\right)$ and $\boldsymbol{\sigma} = \left(\sigma^1,\ldots,\sigma^i,\ldots,\sigma^N\right)$, we could formally define an SSE and the protocol designer's design problem in this work as follows.

**Definition 1 (Stochastically Stable Equilibrium).** A strategy profile $\boldsymbol{\sigma}$ together with a community configuration $\mu$ is an SSE if and only if when $\varepsilon \to 0$

(1) $\boldsymbol{\sigma}$ is the best response of users against $\mu$, i.e. $\sigma^i = \arg\max_{\sigma} v_{\kappa}^{\infty}\left(\sigma^i,\theta^i,\mu\right)$, $\forall i \in \{1,\ldots,N\}$;

(2) $\mu$ is time invariant under the best response dynamics introduced by $\boldsymbol{\sigma}$;

(3) The probability that the community stays at $\mu$ in the long term is larger than with a positive value $\alpha$, where $\alpha$ is determined by the community characteristics.

Therefore, when time goes to infinity, the fraction of time that the community stays at the configuration $\mu$ is bounded away from zero [23]. In contrast to an SSE, an evolutionary stable equilibrium (ESE) [27] satisfies the conditions (1) and (2), i.e. it is stationary when there is no stochastic permutation with $\varepsilon = 0$. Nevertheless, with the existence of stochastic permutations, a community will deviate from an ESE and the fraction of time that the community stays in an ESE might approach 0 in the long term. Hence, an ESE is not necessarily an SSE which is stable against stochastic permutations.

The protocol designer tries to optimize the social welfare $U$ whose maximum value is achieved when all users choose to cooperate with their clients and provide services. This problem can be formulated as $\max_{\kappa} U$. As we will show in the next section, the community converges to configurations belonging to the set of SSE with probability 1 in the long term. Hence, the goal of the protocol designer is to design a social norm $\kappa$ under which users are mutually cooperative in all SSE.

# III. Users' Strategic Learning and The Community's Long-Term Evolution

*A. The MDP formulation and the structure of the optimal policy*

In this section, we analyze the evolution of the community and the design of protocols to induce cooperation among users. We first model a user's learning and adaptation as an MDP, and characterize the structure of a user's optimal policy on strategy adaptation. With this structure, we prove the properties of SSE in Theorem 1, which helps us to design the optimal protocol in Theorem 2 in order to induce users to play cooperatively in the long term.

At the beginning of each period, we assume that each user has a probability $\gamma \in (0,1)$, which we refer to as the adaptation rate, to adapt its service strategy. That is, a user plays its service strategy used in the previous period with probability $1-\gamma$, and adapts to play its best response which satisfies (6) with probability $\gamma$. The adaptation rate does not necessarily to be the same for all users as different users adapt their strategies at different frequencies due to their heterogeneous computation capabilities. Formally, the adaptation of a user could be represented as an optimization problem over its sharing strategy $\sigma \in \Gamma$ in order to maximize its expected long-term utility $v_{\kappa}^{\infty}(\sigma, \theta, \mu)$, given its current knowledge and beliefs on the community which are summarized as follows for the reader's convenience.

**Knowledge**: At the beginning of each period, a user learns its current reputation $\theta$ as well as the current community configuration $\mu$ from the managing device, e.g. the tracker [6]. Hence, a user also learns its current opponent configuration $\eta_{\theta}$.

**Belief**: Each user maintains a belief about the strategies of other users, i.e. their selections in the strategy space $\Gamma$, as described in Section II.D. That is, a user whose reputation is higher than 0 will comply with the social rule $\phi$ with the probability $1-\varepsilon$ in the current period, while a user whose reputation is 0 adopts the service strategy $\sigma_D$ with the probability $1-\varepsilon$.

Given the observations and beliefs, a user's best-response optimization can be formulated as a Markov Decision Process (MDP) [22]. The state in the MDP in each period is a user's current reputation $\theta$ and the opponent configuration $\eta_{\theta}$, which is defined as $s = (\theta, \eta_{\theta}) \in \mathcal{S}$. The action $a = \sigma \in \Gamma$ is represented by a user's serving strategy, which is determined at the beginning of a period based on its state in this period. The one-period reward function $r\left(s,a\right)$ is defined as the expected one-period utility as in (4). Similarly, the long-term reward function $R\left(s\right)$ of a user is defined as the discounted sum $R\left(s\right)=\sum_{t=0}^{\infty}\delta^t r\left(s^{(t)},a^{(t)}\right)$.

[6] This assumption is feasible in online communities such as P2P networks and crowdsourcing applications.

The solution of the MDP is a policy $\pi:\mathcal{S}\rightarrow\Gamma$, which maps each state to a service strategy. The value function is thus defined as the expected long-term utility $v_\kappa^\infty\left(\sigma,\theta,\mu\right)$, under a policy $\pi$, which is formally represented as

$$V^\pi\left(s\right)=E\left\{R\left(s\right)\middle|\pi\right\}=E\left\{r\left(s,a\right)\middle|\pi\right\}+\delta\sum_{s'}p\left(s'\mid s,\pi\left(s\right)\right)V^\pi\left(s'\right), \tag{7}$$

where $p\left(s'\mid s,\pi\left(s\right)\right)$ is the transition probability from $s$ to $s'$ given the action $\pi\left(s\right)$.

The above MDP can be solved using common computation methods such as value iteration and Q-learning [22], with the resulting optimal policy and value function being $\left\{\pi^*\left(s\right)\right\}$ and $\left\{V^*\left(s\right)\right\}$. In any period, the optimal policy instructs a user of reputation $\theta$ who faces an opponent configuration $\eta_\theta$ to play $\pi^*\left(\theta,\eta_\theta\right)$, which is the best response of this user that satisfies (6).

It is to note that if a user chooses, based on its best response, to provide services to a client, this indicates that the user's reputation will be increased by doing so, which leads to an increase future benefit for this user that outweighs the immediate cost it consumes. Hence, if a user provides services to a client of reputation $\tilde{\theta}$ in its best response, it should do the same to clients of any reputation $\tilde{\theta}'>\tilde{\theta}$ in its best response. Because the services to a client of reputation $\tilde{\theta}'>\tilde{\theta}$ results in a higher future benefit and the same immediate cost compared with the services to a client of reputation $\tilde{\theta}$. Consequently, the best response (i.e. the optimal policy in a particular state) should be a threshold-based strategy for any user. This result is formalized in the following lemma.

**Lemma 1.** $\pi^*\left(\theta,\eta\right)$ is a threshold-based strategy as defined in (1) for any $\theta\in\Theta$ and $\eta$.

*Proof*: Omitted due to the space limitation. ■

Therefore, a user adopting $\pi^*\left(\theta,\eta\right)$ will only provide services to clients whose reputations are no less than a threshold and $\pi^*\left(\theta,\eta\right)$ for any $\theta$ and $\eta$ is fully specified by its service threshold. This observation simplifies the above MDP formulation, as we can exclusively consider threshold-based strategies in our analysis and formalize the action in the above MDP as $a\in\mathcal{A}=\left\{0,1,\cdots,L+1\right\}$ and a policy as a mapping $\pi:\mathcal{S}\rightarrow\{0,1,\cdots,L+1\}$. This simplified MDP formulation will be used in the rest of this paper without further notice.

In the rest of this section, we characterize the structure of $\left\{\pi^*\left(s\right)\right\}$, i.e. the optimal service thresholds in the user's best response, which is important for us to understand the evolution of the community that will be analysed in the next section.

First, it can be shown that the optimal service threshold of a user of any reputation $\theta$ in any period is always above the corresponding service threshold $h_\theta$ specified in the social rule, regardless of the opponent configuration $\eta$. That is, $\pi^*(\theta,\eta) \geq h_\theta$ for any $\theta$ and $\eta$. Recall that bad users are referred to those whose reputations are below the social threshold $h$ and good users are referred to those whose reputations are no less than $h$, we discuss the optimal service threshold by analyzing these two groups of users separately. For bad users, the conclusion is obvious, since their optimal service thresholds are always no less than that instructed by the social rule, which is 0. For good users, we prove it by contradiction with the basic idea as follows.

If $\pi^*(\theta,\eta) < h$ for some $\theta \geq h$, it implies that a user of reputation $\theta$ chooses to provide more services to the community than what is required by the social rule. This leads to a higher expected service cost for this user in the current period as well as a higher probability to be punished by the social norm for deviating from $\phi$ [7], and hence a lower expected future utility compared to what it could receive by complying with $\phi$. To sum up, this user will receive a lower expected long-term utility by following $\pi^*(\theta,\eta)$ than what it could receive by complying $\phi$, which contradicts the fact that $\pi^*(\theta,\eta)$ is the best response that satisfies (6). Since the argument is valid for all $\theta \geq h$, we have the conclusion.

**Lemma 2.** $\pi^*(\theta,\eta) \geq h_\theta$, for any $\theta \in \Theta$ and $\eta$.

*Proof*: See Appendix A. ■

Lemma 2 proves that a user can never benefit by choosing a strategy in its best response which has a lower service threshold than that specified in the social rule. Hence, no user will voluntarily provide more services than what is required by the social norm.

Following a similar idea, we prove the following lemma, which shows the monotonicity on the service thresholds in bad (good) users' best responses.

**Lemma 3.** $\pi^*(\theta_1,\eta) \geq \pi^*(\theta_2,\eta)$ if $\theta_1 < \theta_2 < h$, and $\pi^*(\theta_1,\eta) \geq \pi^*(\theta_2,\eta)$ for any $h \leq \theta_1 < \theta_2$.

*Proof*: Omitted here due to the space limitation. ■

Lemma 3 states that a bad (good) user with a low reputation will never provide more services than a bad (good) user with a high reputation, when they are confronted with the same opponent configuration. That is, the optimal service threshold in a bad (good) user's best response (weakly) decreases against its reputation.

We use bad users as the example to briefly explain this lemma. From (4), it can be determined that the expected amount of services that has to be provided by a bad user in one period only depends on its current opponent configuration and its selection of the sharing strategy, but is independent on its current reputation. Since all users in the community choose threshold-based strategies in their best responses as proved in Lemma 1, each bad user can expect to receive the same average amount of service which is independent on its reputation and the service strategy it chooses.

[7] It should be noted that the social norm does not only punish users who do not provide services as required. If a user provides service to another user who is supposed not to receive services by the social norm, this user itself will also be punished.

Suppose we have $\pi^*\left(\theta+1,\eta\right) > \pi^*\left(\theta,\eta\right)$ for some $\theta < h-1$. If a bad user $i$ of reputation $\theta$ switches its service threshold from $\pi^*\left(\theta,\eta\right)$ to $\pi^*\left(\theta+1,\eta\right)$, its expected one-period utility increases by $\Delta r\left(\theta\right)$ since user $i$ saves some immediate service cost with a higher service threshold. However, due to the optimality of $\pi^*$, the expected long-term utility of user $i$ will decrease. In other words, user $i$'s loss on its future utility, which is $\delta\sum_{\eta'} p\left(\eta' \mid \eta\right)\left[p\left(0 \mid \theta,\eta,\pi^*\left(\theta+1,\eta\right)\right) - p\left(0 \mid \theta,\eta,\pi^*\left(\theta,\eta\right)\right)\right]\left[V^*\left(\theta+1,\eta'\right) - V^*\left(0,\eta'\right)\right]$ outweighs $\Delta r\left(\theta\right)$. Now consider another bad user $j$ of reputation $\theta+1$. If it switches its service threshold from $\pi^*\left(\theta+1,\eta\right)$ to $\pi^*\left(\theta,\eta\right)$, its expected one-period utility decreases by $\Delta r\left(\theta\right)$ while its future utility is increased by $\delta\sum_{\eta'} p\left(\eta' \mid \eta\right)\left[p\left(0 \mid \theta+1,\eta,\pi^*\left(\theta+1,\eta\right)\right) - p\left(0 \mid \theta+1,\eta,\pi^*\left(\theta,\eta\right)\right)\right]\left[V^*\left(\theta+2,\eta'\right) - V^*\left(0,\eta'\right)\right]$.

Since we have proved in Lemma 2 that $V^*\left(\theta+2,\eta'\right) \geq V^*\left(\theta+1,\eta'\right)$ always holds for all $\theta$ and $\eta'$ and $p\left(0 \mid \theta+1,\eta,\pi^*\left(\theta+1,\eta\right)\right) - p\left(0 \mid \theta+1,\eta,\pi^*\left(\theta,\eta\right)\right) = p\left(0 \mid \theta,\eta,\pi^*\left(\theta+1,\eta\right)\right) - p\left(0 \mid \theta,\eta,\pi^*\left(\theta,\eta\right)\right)$, we have user $j$'s increase on its future utility being higher than $\Delta r\left(\theta\right)$ and hence, user $j$ monotonically increases its expected long-term utility by choosing $\pi^*\left(\theta,\eta\right)$. This contradicts the fact that $\pi^*\left(\theta+1,\eta\right)$ is user $j$'s optimal service threshold and hence, $\pi^*\left(\theta+1,\eta\right) > \pi^*\left(\theta,\eta\right)$ for some $\theta < h-1$ does not hold. A similar analysis can be applied to the optimal service thresholds of good users.

As a result of Lemma 2 and 3, a user will always try to learn to exploit the social norm and to provide fewer services than what is required by the social rule. Hence, a user's best response adaptation will not increase the social welfare of a community, and in most cases, will decrease it.

*B. The community's long-term evolution and the design of the optimal protocol*

In this section, we examine the evolution of the community under the best response dynamics. We define the strategy configuration of the community as $\left\{\pi_\mu^*\left(\theta\right)\right\}_{\theta=0}^{L}$. Particularly, $\pi_\mu^*\left(\theta\right)$ represents the optimal service threshold that a user of reputation $\theta$ will choose in its best response when the community configuration is $\mu$. For $\mu = \left\{n\left(\theta\right)\right\}_{\theta=0}^{L}$ and $\eta = \left\{m\left(\theta\right)\right\}_{\theta=0}^{L}$, we have $\pi_\mu^*\left(\theta'\right) = \pi^*\left(\theta',\eta\right)$ for all $\theta'$ if $m\left(\theta\right) = n\left(\theta\right)$ for all $\theta \neq \theta'$ and $m\left(\theta'\right) = n\left(\theta'\right) - 1$.

A Markov chain analysis is employed to analyze the community configuration $\mu$'s evolution. For notational convenience, let $Hist$ denote the history of the community, which includes past community configurations and the users' actions until the current period.

Given the social norm $\kappa = (\phi, \tau)$, it is already shown that the best response of a user is fully determined by its belief and its opponent configuration, or alternatively, the user's reputation and the community configuration. Meanwhile, the probability of the community configuration in the next period being $\mu'$, denoted as $p(\mu' \mid Hist)$, is also determined by $\mu$ and $\left\{\pi_\mu^*(\theta)\right\}_{\theta=0}^{L}$. As a result, we have $p(\mu' \mid Hist) = p\left(\mu' \mid \mu, \left\{\pi_\mu^*(\theta)\right\}_{\theta=0}^{L}\right) = p(\mu' \mid \mu)$, which implies that the community configuration evolves as a Markov chain on the finite space

$$\mathcal{U} = \begin{Bmatrix} \mu = (n(0), n(1), \ldots, n(L)) \\ \mid n(\theta) \in \mathbb{N} \ for \ 0 \le \theta \le L, and \ \sum_{\theta=0}^{L} n(\theta) = N \end{Bmatrix} \tag{8}$$

whose size is $|\mathcal{U}| = \begin{pmatrix} N+L \\ L \end{pmatrix}$. The transition probabilities are given by $p_\varepsilon(\mu' \mid \mu)$ between any two configurations $\mu$, $\mu' \in \mathcal{U}$, which has the subscript $\varepsilon$ to highlight their dependences on the operation error. $P_\varepsilon = \left[p_\varepsilon(\mu' \mid \mu)\right]$ is the transition matrix. Note that with $\varepsilon > 0$, this Markov chain is communicating. It is well-known then that this Markov chain is irreducible and aperiodic, which introduces a unique stationary distribution. Indexing all configurations in $\mathcal{U}$ from 0 to $|\mathcal{U}| - 1$ and letting $\Delta_{|\mathcal{U}|} \equiv \left\{\boldsymbol{q} \in \mathbb{R}^{|\mathcal{U}|} \mid q(i) \ge 0 \ for \ i = 0,1,2,\cdots,|\mathcal{U}|\text{-}1 \ and \ \sum_i q(i) = 1\right\}$ be the $|\mathcal{U}|$-dimensional simplex, any $\boldsymbol{q} \in \Delta_{|\mathcal{U}|}$ then represents a probability distribution on the set of community configurations with $q(i)$ being the frequency that the community is in the $i$-th configuration in the long term. A stationary distribution is a row vector $\boldsymbol{\omega}_\varepsilon = \left(\omega_\varepsilon(0), \omega_\varepsilon(1), \ldots, \omega_\varepsilon(|\mathcal{U}| - 1)\right) \in \Delta_{|\mathcal{U}|}$ satisfying $\boldsymbol{\omega}_\varepsilon = \boldsymbol{\omega}_\varepsilon P_\varepsilon$. When $P_\varepsilon$ is strictly positive, not only $\boldsymbol{\omega}_\varepsilon$ exists and is unique, it also preserves the following important properties.

**Lemma 4.** $\boldsymbol{\omega}_\varepsilon$ preserves the following properties:

(1) Stability: Starting from an arbitrary initial configuration distribution $\boldsymbol{q}$, we have that

$$\lim_{t \to \infty} \boldsymbol{q} P_\varepsilon^t \to \boldsymbol{\omega}_\varepsilon . \tag{9}$$

(2) Ergodicity: Starting from an arbitrary initial configuration $\mu$, the fraction of time that the community stays in $i$-th configuration in the long term is $\boldsymbol{\omega}_\varepsilon(i)$, for any $0 \le i \le |\mathcal{U}| - 1$. ■

Next, we analyse the stochastic stability of the community when $\varepsilon \to 0$. To facilitate the analysis, we first define the concept of the limiting configuration distribution.

**Definition 2 (Limiting Configuration Distribution).** The limiting configuration distribution of the community is defined by

$$\bar{\boldsymbol{\omega}} = \lim_{\varepsilon \to 0} \boldsymbol{\omega}_{\varepsilon}. \tag{10}$$

The existence and the uniqueness $\bar{\boldsymbol{\omega}}$ can be proved as in [8].

**Lemma 5.** There exists a unique limiting distribution for the community configurations.

*Proof*: Proved in [8]. ■

Correspondingly, we define the concept of stochastically stable configuration as follows.

**Definition 3 (Stochastically Stable Configuration).** A community configuration $\mu$ with index $i$ is called a *stochastically stable configuration* if and only if $\bar{\omega}\left(i\right) > 0$.

According to Definition 1 and 3, it can be concluded that a community configuration belonging to an SSE is always a stochastically stable configuration. In the following theorem, it can be further proved in another direction that a stochastically stable configuration always belongs to one of the SSE. Hence, the set of stochastically stable configurations and the set of configurations that belong to SSE contain each other and are thus equivalent.

The following theorem also proves that a stochastically stable configuration should be composed exclusively by user populations of reputations 0 and $L$. Alternatively speaking, users in a SSE should either always follow $\sigma_D$ to provide no service to any other user and maintain a reputation of 0, or always comply with the social rule $\phi$ and maintain a reputation of $L$.

**Theorem 1.** A community configuration $\mu = \left\{n\left(0\right), \cdots, n\left(L\right)\right\}$ belongs to an SSE if and only if it is a stochastically stable configuration. $\mu$ also preserves the following property

$$n\left(\theta\right) = 0, \text{ for all } \theta \in \{1, \cdots L-1\}. \tag{11}$$

*Proof*: See Appendix B. ■

According to Theorem 1, there are $N+1$ community configurations that satisfy (11) and could possibly be stochastically stable configurations. With the slight abuse of notations, we use $\mu_k$ to represent the community configuration with $n\left(L\right) = k$ and $n\left(0\right) = N-k$ for $k \in \left\{0, \ldots N\right\}$. Since users of reputation 0 always play $\sigma_D$ in an SSE, users are not mutually cooperative in any configuration $\mu_k$ for $k \in \left\{0, \ldots N-1\right\}$. Therefore, the protocol designer should design a social norm $\kappa$ under which $\mu_N$ is the unique stochastically stable configuration in the community, while all users are complying with the social rule and are mutually cooperative with each other. The design problem of the optimal protocol stated in Section II.D is reformulated as follows:

**Definition 4 (Protocol Designer's Problem).**

$$\begin{array}{ll} select & \kappa \\ s.t. & \mu_N \text{ is the unique stochastically stable configuration} \end{array}. \quad (12)$$

The design problem (12) is solved in the following theorem, where we determine the sufficient and necessary conditions for $\mu_N$ to be the unique stochastically stable configuration by examining the *basin of attraction* of each community configuration [8]. A configuration is a stochastically stable configuration if and only if it has the largest basin of attraction among all community configurations.

**Theorem 2.** The community converges to a unique stochastically stable configuration $\mu_N$ in the long term as $n\left(L\right) = N$ and $n\left(\theta\right) = 0$ for all $\theta < L$ if and only if $\delta > \frac{c}{b}$ and $h < H$, where $H$ is the solution of the following equation

$$\delta^H b - \delta^{H-1} c = \left(1 - \delta^H\right) cH . \quad (13)$$

*Proof*: See Appendix C. ■

Theorem 2 characterizes the limiting behavior of the community after sufficiently long time. It can be used as a guideline for a protocol designer to regulate the online community. For a better illustration, we plot the change of $H$ against the discount factor $\delta$ and the service cost to benefit ratio $c / b$ in Figure 2 (a), and Figure 2 (b) shows the region (marked in blue) of $\delta$ and $c / b$ where a feasible social threshold $h$, i.e. $h \geq 1$, can be found to induce the community converging to $\mu_N$. According to Figure 2, Theorem 2 provides three sufficient conditions for constructing a mutually cooperative community under stochastic permutations: (1) the service cost should be sufficiently low to ensure a small service cost to benefit ratio $c / b$; (2) users are sufficiently patient; (3) the social rule provides sufficient services to users with a low service threshold.

We briefly explain the above three conditions. Condition (1) is intuitive. By deviating from the social norm, the maximum gain on a user's one-period utility is the saving of its immediate service cost, which is proportional to $c$, at the loss of its future utility which is proportional to $b$. Hence, when $c / b$ decreases, the current gain of a user also decreases with its future loss increasing at the same time, which provides stronger incentives for a user to comply with the social rule. A similar analysis can be applied to Condition (2). The discount factor $\delta$ adjusts the weights that a user places on its current and future utilities. With a larger $\delta$, a user puts a higher weight on its future utility, and thus becomes more interested in increasing its reputation and obtaining a higher future utility rather than deviating to save its immediate service cost. As a result, the incentive for a user to comply with the social rule increases. Condition (3) contradicts the traditional opinion that a user's incentive will increase when the punishment in the protocol becomes more severe. In our framework, this statement is correct only for good users. As outlined by the proof, when the punishment is too severe, which is represented here by a high social threshold in the social rule, a bad user has to wait a long period of time to recover its reputation (i.e. becoming a good user), which harms its incentive to comply with the social rule. This prohibits the protocol designer from increasing $h$ arbitrarily.

## IV. EXPERIMENTS

In our experiments, we illustrate the users' adaptation behavior and the impact of the proposed social norm based protocols on P2P file-sharing communities. We assume that a number of $N = 1000$ users are deployed in the community. All users have the same download rate of 1Mbps. In each experiment, users exchange a single file of approximate size 100 Mbits, which is divided into chunks of equal sizes. We use the chunk size to measure the download benefit and the upload cost in a transaction. A larger size results in a higher benefit and a higher cost. Since each chunk has the same size, each successful transaction (i.e. the server uploads the requested chunk to the client) incurs a constant benefit to the client and a constant cost to the server. We deploy a reputation set $\Theta = \{0,1,2,3\}$, i.e. $L = 3$, and $\Theta$ is kept fixed throughout the experiments.

In the first experiment, we run the community for $10^8$ periods to measure its long-term evolution. After every $10^7$ periods, we sample the reputation distribution in the community at that moment. The results are illustrated in Figure 3. Figure 3 (a) plots how the reputation distribution varies over time when $\varepsilon = 0.2$. As it shows, the reputation distribution oscillates and cannot converge to a unique stationary point. Figure 3 (b) plots the evolution of the community when $\varepsilon = 0.05$. Since $\varepsilon$ is sufficiently small and close to 0, the operation error happens rarely. As a result, the community converges to the stochastically stable configuration as proved by Lemma 5. After $6\times 10^7$ periods, about 90% of users are of reputations $L$ in each period and 10% of users are of reputation 0, while the fractions for users of reputations other than 0 and $L$ almost diminish.

In the remainder of the experiments, we focus on the stochastic stability of the community by considering a small operation error probability $\varepsilon = 0.05$.

We first analyze how different community characteristics influence the evolution of the community and the stochastically stable configuration that the community finally converges to in the long term. Figure 4 plots how the fraction of users of reputation $L$, i.e. $n(L)$, after $10^8$ periods changes against the discount factor $\delta$. It also demonstrates how the service benefit $b$ as well as the social threshold $h$ impacts the resulting stochastically stable configuration. As what can be seen from this figure, $n(L)$ in the long term monotonically increases with $\delta$, which is consistent with Theorem 2 since users become more patient during their adaptations with $\delta$ becoming larger and hence more users choosing to mutually cooperate in the long term. Compared to the scenario when $b = 5$, a user's saving on its immediate service cost outweighs its future benefit for obtaining a high reputation when $b = 3$, and hence, $n(L)$ is smaller in this case. Similar results are obtained for the social threshold $h$, as users becoming easier to be incentivized to comply with the social rule and gain reputation $L$ when $h$ is small.

All the analysis and experiments so far are conducted in a community where all users are homogeneous and have the same discount factor $\delta$, which we refer to as a pure community. However, in practice, different users should have different levels of patience and thus different discount factors, which form a so-called "mixed" community. In the next experiment, we analyze the evolution of a community with two groups of users. Each group contains $N = 1000$ users. Users from Group 1 have a discount factor of 0.3 and users from Group 2 have a discount factor of 0.6. The discount factor is private information to each user and is not revealed to others. Since the convergence analysis in Theorem 1 does not depend on the users' discount factors, its result still holds. Therefore, the community also converges to an SSE which satisfies (11). Figure 5 plots the average reputation distribution in the community after $10^8$ periods [8].

The white and grey bars in Figure 5 show the configurations that the community converges to when it is purely composed of users with $\delta = 0.6$ and of users with $\delta = 0.3$, respectively. 28% of users converges to the reputation 0 and 72% of users converges to the reputation $L$ in the pure community with $\delta = 0.6$, while 82% of users converges to the reputation 0 and 18% of users converges to the reputation $L$ in the pure community with $\delta = 0.3$ since users are less patient here. The red and black bars jointly represent the reputation distribution in the mixed community where users of two different discount factors co-exist. In general, the number of users in the mixed community choosing to play $\sigma_D$ and maintain the reputation 0 is more than that in the pure community with $\delta = 0.3$ and fewer than that in the pure community with $\delta = 0.6$. However, among users with $\delta = 0.3$, 60% of them choose to play $\sigma_D$ in the long term, which is lower than the percentage in the pure community in which they only interact with each other. On the contrary, 34% of users with $\delta = 0.6$ choose to play $\sigma_D$ in the long term, which is higher than that in the corresponding pure community. This phenomenon occurs because a user with $\delta = 0.3$ can expect a higher future utility and hence have a larger incentive to comply with the social rule in a mixed community than in a pure community, since users with $\delta = 0.6$ are more patient and willing to provide services. This also explains the decrease of good users in the population of users with $\delta = 0.6$, as they observe a larger population of users playing $\sigma_D$, which reduces their expectation for the future utility and hence the incentives to contribute services in lieu of better reputations.

In the following experiment, we assume that the community characteristics, e.g. the utility structure $(b, c)$ and the discount factor $\delta$ are not only heterogeneous among users but also vary over time. As an example, we use $b$ as the representative variable to plot the result. Each user's evaluation on its service benefit varies over time following a Gaussian distribution with the mean $\overline{b} = 3$ and the variance $\sigma^2 = 0.01$. Figure 6 depicts the social welfare of the community over time. We consider two selections on the social threshold as $h = 1$ and $h = 2$. In both cases, the social welfare with $b$ being variable (solid lines) is smaller than the social welfare with $b$ being constant (dotted lines). This is due to the fact that the designed protocol only provides users sufficient incentives to comply with the social norm when $b$ is at its mean value 3. When $b$ deviates from its mean value, users may have incentives to adapt to the best response and deviate from the social norm. In addition, since most users maintain reputation $L$ in the stochastically stable equilibrium when $\overline{b} = 3$ and $\delta = 0.5$, the social norm with $h = 2$ can provide stronger incentives for users of reputation $L$ to follow the social rule because it providers larger incentives for users of reputation $L$. Therefore, it is more robust against the variation on $b$, which maintains $n(L)$ at a higher level than the social norm with $h=1$. As a result, the social norm with $h=2$ delivers higher social welfare for both $N=500$ and $N=1000$ when $b$ is variable.

[8] Since the fraction of users of reputations other than 0 and $L$ are negligible, we focus on users of reputations 0 and $L$.

In the final part of the experiments, instead of employing the fixed belief model as in Section II.D, we simulate the evolution of the community when each user constructs a belief on the service strategies of other users based on its past interactions. In particular, each user maintains a belief matrix $O$ of size $(L+1)\times(L+2)$. An element $O(\theta,l)$ represents the probability that a user believes that its opponent of reputation $\theta$ will choose the service strategy of threshold $l$, hence $\sum_{l=0}^{L+1} O(\theta,l)=1$ for all $\theta$.

In the beginning of the experiment, each user believes that all users other than itself will comply with the social rule. Hence, the belief matrix is initialized with $O(\theta,0)=1$ for any $\theta<h$ and $O(\theta,h)=1$ for any $\theta\geq h$. After the $t$-th transaction, if a user of reputation $\theta'$ is served by another user of reputation $\theta$, it updates its belief matrix by increasing the weights on the entries $\left\{O(\theta,l)\right\}_{l=0}^{\theta'}$, which indicates its increased belief on the fact that the users of reputation $\theta$ has a service threshold no higher than $\theta'$. On the other hand, if the user's service request is rejected by another user of reputation $\theta$, it believes that the users of reputation $\theta$ has a service threshold higher than $\theta'$ and hence increases the weights on the entries $\left\{O(\theta,l)\right\}_{l=\theta'+1}^{L+1}$. The belief update follows the rule below, where $z$ is the level of contribution that a user observes from its server:

$$O^{(t)}(\theta,l)=\begin{cases}\dfrac{O^{(t-1)}(\theta,l)\times(t-1)+\dfrac{1}{\theta'+1}z}{t}\\[2ex] \dfrac{O^{(t-1)}(\theta,l)\times(t-1)+\dfrac{1}{L+1-\theta'}(1-z)}{t}\end{cases}, \tag{14}$$

Following this belief updating rule, we run the experiment for $10^8$ periods to measure the long-term evolution of the community. To compare with the result in Figure 3 (b), we use the same parameters as $\varepsilon=0.05$, $b=3$, $c=1$ and $\delta=0.5$. Figure 7 plots the results when the community population $N=500$ and 1000, respectively. In both cases, the community configuration converges with users employing the belief updating rule (14). Since the stochastically stable configuration is simply determined by the community characteristics, the community converges to the same configuration as in Figure 3 (b) where a fixed belief is employed. However, due to the belief updating, it takes a longer period of time for the community to converge to the stochastically stable configuration. In particular, the convergence time is $7\times10^7$ periods for the community with $N=500$ and $9\times10^7$ periods for the community with $N=1000$, which also reflects the fact that users are easier to get their opinions and strategies converge to the common sense in small communities.

# V. CONCLUSION AND FUTURE RESEARCH

We have studied the problem of designing social norm based protocols for online communities and analyzed the learning behavior of users under such protocols. Knowledge on the evolution of the community in the long term can facilitate the protocol designers to design protocols which achieve efficient social welfare. Our framework can be extended in several directions, among which we mention four. First, users in the community do not necessarily need to be homogeneous as discussed in Section IV. Different users can have different benefits and costs for the service received/provided. Also, they can choose different discount factors $\delta$ when evaluating the long-term utility. The discount factor that a user chooses can be dynamically adjusted over time depending on its own expected lifetime in the community. Some preliminary simulation results have been illustrated in Section IV, but a more extensive analysis can be performed to analyze how the user heterogeneity impacts the design of efficient protocols. Second, clients can use more complicated decision rules while reporting the servers' actions to the community manager in order to maximize their own long-term utility, instead of always reporting truthfully. Third, online communities may be subject to practical constraints such as topological constraints, in which users can only observe the local information and different users at different locations do not necessarily share the same community information. Hence, the analysis in this paper needs to be extended to scenarios where users adapt based on partial and heterogeneous information. Fourth, users adopt a simple belief model in this paper. However, a more sophisticated belief model can be introduced into our framework such that users can update their beliefs on others based on their observation. For example, the formation of user beliefs and opinions in social networks are extensively studied in [29] and [30]. Understanding how the evolutions of users' beliefs and users' strategies will impact each other will also form an important future research direction.

## APPENDIX A

## PROOF OF LEMMA 2

Since $h_\theta = 0$ for $\theta < h$ and $\pi^*\left(\theta, \eta\right) \geq 0$ for any $\theta$, the conclusion holds for $\theta < h$. We then analyze the case with $\theta \geq h$. We first prove that $V^*\left(\theta_1, \eta\right) \leq V^*\left(\theta_2, \eta\right)$ for any $\theta_1 < \theta_2$, i.e. by playing the best response, a user of a higher reputation will not receive a lower expected long-term utility than that is received by a user of a lower reputation. It is obvious that a good user's optimal value function is always higher than that of a bad user, due to the extra services it receives from other good users. Therefore, we only have to prove that $V^*\left(\theta, \eta\right) \leq V^*\left(\theta + 1, \eta\right)$ holds when $\theta < h - 1$ and $\theta \geq h$, respectively. Due to the similar logic, we only show the later part in our proof, i.e. $V^*\left(\theta, \eta\right) \leq V^*\left(\theta + 1, \eta\right)$ holds when $\theta \geq h$.

Now suppose the set of optimal value functions for good users is not in the increasing order with $\theta$. Denote $\overline{\theta}$ as a reputation which preserves the property that $\overline{\theta} = \max\left\{\theta \mid V^*\left(\theta, \eta\right) > V^*\left(\theta + 1, \eta\right), \exists \eta\right\}$. It is known that the optimal value functions preserve the stationary property as

$$\begin{aligned}
V^*\left(\overline{\theta},\eta\right) &= r\left(\overline{\theta},\eta,\pi^*\left(\overline{\theta},\eta\right)\right) + \delta\sum_{\eta'} p\left(\eta' \mid \eta\right) p\left(0 \mid \overline{\theta},\eta,\pi^*\left(\overline{\theta},\eta\right)\right) V^*\left(0,\eta'\right) \\
&\quad + \delta\sum_{\eta'} p\left(\eta' \mid \eta\right)\left[1 - p\left(0 \mid \overline{\theta},\eta,\pi^*\left(\overline{\theta},\eta\right)\right)\right] V^*\left(\overline{\theta}+1,\eta'\right) \\
V^*\left(\overline{\theta}+1,\eta\right) &= r\left(\overline{\theta}+1,\eta,\pi^*\left(\overline{\theta}+1,\eta\right)\right) + \delta\sum_{\eta'} p\left(\eta' \mid \eta\right) p\left(0 \mid \overline{\theta}+1,\eta,\pi^*\left(\overline{\theta}+1,\eta\right)\right) V^*\left(0,\eta'\right) \\
&\quad + \delta\sum_{\eta'} p\left(\eta' \mid \eta\right)\left(1 - p\left(0 \mid \overline{\theta}+1,\eta,\pi^*\left(\overline{\theta}+1,\eta\right)\right)\right) V^*\left(\min\left\{\overline{\theta}+2, L\right\},\eta'\right)
\end{aligned} \quad . (15)$$

Here we consider a new policy $\tilde{\pi}$ with $\tilde{\pi}\left(\overline{\theta}+1,\eta\right) = \pi^*\left(\overline{\theta},\eta\right)$ and $\tilde{\pi}\left(\theta,\eta'\right) = \pi^*\left(\theta,\eta'\right)$ for all $\theta \neq \overline{\theta}+1$ or $\eta \neq \eta'$. Substituting this into (15), the resulting value function for $\overline{\theta}+1$ now becomes

$$\begin{aligned}
\tilde{V}\left(\overline{\theta}+1,\eta\right) &= r\left(\overline{\theta}+1,\eta,\tilde{\pi}\left(\overline{\theta}+1,\eta\right)\right) + \delta\sum_{\eta'} p\left(\eta' \mid \eta\right) p\left(0 \mid \overline{\theta},\eta,\tilde{\pi}\left(\overline{\theta},\eta\right)\right) V^*\left(0,\eta'\right) \\
&\quad + \delta\sum_{\eta'} p\left(\eta' \mid \eta\right)\left[1 - p\left(0 \mid \overline{\theta},\eta,\tilde{\pi}\left(\overline{\theta},\eta\right)\right)\right] V^*\left(\min\left\{\overline{\theta}+2, L\right\},\eta'\right)
\end{aligned} \quad . \quad (16)$$

Comparing $V^*\left(\overline{\theta},\eta\right)$ and $\tilde{V}\left(\overline{\theta}+1,\eta\right)$, we have that

$$\begin{aligned}
& r\left(\overline{\theta},\eta,\pi^*\left(\overline{\theta},\eta\right)\right) \leq r\left(\overline{\theta}+1,\eta,\tilde{\pi}\left(\overline{\theta}+1,\eta\right)\right) \\
& p\left(0 \mid \overline{\theta},\eta,\pi^*\left(\overline{\theta},\eta\right)\right) = p\left(0 \mid \overline{\theta}+1,\eta,\tilde{\pi}\left(\overline{\theta}+1,\eta\right)\right). \\
& V^*\left(\overline{\theta}+1,\eta'\right) \leq V^*\left(\min\left\{\overline{\theta}+2, L\right\},\eta'\right)
\end{aligned} \quad (17)$$

Hence, $\tilde{V}\left(\overline{\theta}+1,\eta\right) \geq V^*\left(\overline{\theta},\eta\right) > V^*\left(\overline{\theta}+1,\eta\right)$, which violates the stationary principle of value iteration. Therefore, $\pi^*\left(\overline{\theta}+1,\eta\right)$ is not the optimal policy and can be further improved. Consequently, we have that $\left\{V^*\left(\theta,\eta\right)\right\}_{\theta=h}^{L}$ should be monotonically (weakly) increasing against $\theta$ for any $\eta$.

Assuming that there is $\pi^*\left(\theta,\eta\right) < h$ for some $\theta \geq h$ and $\eta$, we compare two choices of service thresholds $a_1 = \pi^*\left(\theta,\eta\right)$ and $a_2 = h$. Based on the property of the optimal value function, $V^*\left(\theta,\eta\right)$ should not be further improved by selecting any action other than $a_1$. Nevertheless, if we choose $a_2$ as the service threshold for users of reputation $\theta$, then it is obvious that $r\left(\theta,\eta,a_1\right) \leq r\left(\theta,\eta,a_2\right)$. Moreover, we have that $p\left(0 \mid \theta,\eta,a_1\right) > p\left(0 \mid \theta,\eta,a_2\right)$ as $a_1$ deviates from the social norm. Since $V^*\left(\theta+1,\eta\right) \geq V^*\left(0,\eta\right)$, we also conclude that

$$\begin{aligned}
& \delta\left(1 - p\left(0 \mid \theta,\eta,a_1\right)\right) V^*\left(\theta+1,\eta\right) + \delta p\left(0 \mid \theta,\eta,a_1\right) V^*\left(0,\eta\right) \\
& \leq \delta\left(1 - p\left(0 \mid \theta,\eta,a_2\right)\right) V^*\left(\theta+1,\eta\right) + \delta p\left(0 \mid \theta,\eta,a_2\right) V^*\left(0,\eta\right)
\end{aligned} . \quad (18)$$

Consequently, choosing $a_2$ guarantees a long-term reward which is at least as good as $a_1$ for users of reputation $\theta$. Finally, upon the consideration that the optimal policy should be unique, we can determine that any $a < h$ should not be optimal and hence $\pi^*\left(\theta,\eta\right) \geq h$. ■

## APPENDIX B

## PROOF OF THEOREM 1

To prove this, we first need to define the concept of an absorbing configuration and an absorbing class of configurations.

**Definition 5 (Absorbing Configuration).** An absorbing configuration $\mu$ preserves the property that $\lim_{\varepsilon\to 0} p_\varepsilon\left(\mu \mid \mu\right) = 1$.

**Definition 6 (Absorbing Class of Configurations).** An absorbing class of configurations $M$ preserves the following property that for any two configurations $\mu \in M$ and $\mu' \notin M$, the transition probability $\lim_{\varepsilon\to 0} p_\varepsilon\left(\mu' \mid \mu\right) = 0$.

It is obvious that an absorbing configuration is also a degenerated absorbing class of configurations. Correspondingly, we could also define the concept of an irreducible absorbing class of configurations.

**Definition 7 (Irreducible Absorbing Class of Configurations).** An irreducible absorbing class of configurations is an absorbing class which is not the super set of any other absorbing class.

Since the limiting distribution $\bar{\omega}$ is dealing with the case when $\varepsilon \to 0$, it can be concluded that a stochastically stable configuration should also be an absorbing configuration or within an irreducible absorbing class of configurations. In the rest of this proof, we first show that there is at least one community configuration which is absorbing. We then show that there is no irreducible absorbing class of configurations that does not include absorbing configurations. Hence, we prove that each stochastically stable configuration is an absorbing configuration.

### *A. The existence of absorbing configurations*

Consider a configuration $\mu = \left\{n(0),\cdots,n(L)\right\}$ which does not satisfy (11), i.e. there exists positive fraction of users in $\mu$ of reputations other than 0 and $L$. After one period, there is always a positive probability for $\mu$ to move to some other configuration. Hence, $\mu$ is not an absorbing configuration.

Next let us consider a configuration $\mu = \left\{n\left(0\right),\cdots,n\left(L\right)\right\}$ which satisfies (11), i.e. $n\left(0\right) + n\left(L\right) = N$. To make this configuration absorbing, we have to ensure that users of reputation 0 choose $\sigma_D$ in the best response and users of reputation $L$ choose to comply with $\phi$ in the best response.

#### *1) Enforcing $\sigma_D$ on users of reputation 0*

In this section, we first discuss the condition for users of reputation 0 to choose $\sigma_D$ in their best responses.

For a user $i$ of reputation 0, it confronts an opponent configuration $\eta_0 = \{m_0(0), \cdots, m_0(L)\}$, with $m_0(0) = n(0) - 1$ and $m_0(L) = n(L) = N - 1 - m_0(0)$. For notation convenience, we also use $p_0(0) = \frac{n(0)-1}{N-1}$ and $p_0(L) = \frac{n(L)}{N-1}$ to represent the fractions of users in user $i$'s opponent configuration. We only have to consider $\{\pi^*(\theta, \eta_0)\}_{\theta=0}^{L}$ in our analysis since user $i$ maintains a belief that all users will follow their current strategies and thus its opponent configuration remains unchanged over time [9]. Therefore, we have the optimal value functions for user $i$ to be

$$V^*(\theta, \eta_0) = r(\theta, \eta_0, \pi^*(\theta, \eta_0)) + \delta p(0 \mid \theta, \eta_0, \pi^*(\theta, \eta_0)) V^*(0, \eta_0) + \delta[1 - p(0 \mid \theta, \eta, \pi^*(\theta, \eta_0))] V^*(\theta + 1, \eta_0). \tag{19}$$

Since there are only users of reputation 0 and users of reputation $L$ in $\eta_0$, only three choices of actions exist for $\pi^*(\theta, \eta_0)$ of any $\theta \le h - 1$ [10]: $\pi^*(\theta, \eta_0) = 0$ implying to provide services to all users; $\pi^*(\theta, \eta_0) = h$ implying to provide services only to good users; and $\pi^*(\theta, \eta_0) = L + 1$ implying not to provide services to any user. As it has been proved that $\pi^*(\theta, \eta) \ge h$ for any $\theta \ge h$, only two choices of actions exist as the optimal service threshold for any $\theta \ge h$: $\pi^*(\theta, \eta_0) = h$ and $\pi^*(\theta, \eta_0) = L + 1$.

Lemma 2 has proved that $\pi^*(\theta, \eta_0) \ge \pi^*(\theta + 1, \eta_0)$ for any $\theta \ge h$. Recursively, it is easy to show that $\pi^*(\theta, \eta_0) = \pi^*(\theta + 1, \eta_0)$ and thus $V^*(\theta, \eta_0) = V^*(\theta + 1, \eta_0)$ for all $\theta \ge h$ are always true. Hence in the rest of this proof, we use $V^*(h, \eta_0)$ to represent any $V^*(\theta, \eta_0)$ with $\theta \ge h$ for convenience.

Regarding the fact that the optimal service threshold (weakly) decreases against the reputation $\theta$ when $\theta < h$, there exists two integers $\underline{k}, \overline{k} \in \{-1, 0, \ldots, h-1\}$ with $\underline{k} \le \overline{k}$. User $i$'s optimal policy can be characterized as: $\pi^*(\theta, \eta_0) = L + 1$ when $0 \le \theta \le \underline{k}$, $\pi^*(\theta, \eta_0) = h$ when $\underline{k} + 1 \le \theta \le \overline{k}$, $\pi^*(\theta, \eta_0) = 0$ when $\overline{k} + 1 \le \theta \le h - 1$. We can thus rewrite the optimal value function of user $i$ as follows:

$$V^*(\theta, \eta_0) = \begin{cases} 0, & if\ 0 \le \theta \le \underline{k} \\ p_0(L)[-c + \delta V^*(\theta + 1, \eta_0)], & if\ \underline{k} + 1 \le \theta \le \overline{k} \\ [-c + \delta V^*(\theta + 1, \eta_0)], & if\ \overline{k} + 1 \le \theta \le h - 1 \end{cases}. \tag{20}$$

---

[9] It should be noted that since $\varepsilon \to 0$, each user maintains a belief that any other user will follow its previous service strategy with probability 1.

[10] Any other strategy has the same effect with these three strategies in realization.

Now we consider $V^*\left(\theta,\eta_0\right)$ for any $\underline{k}+1\le\theta\le\overline{k}$. If $-c+\delta V^*\left(\theta+1,\eta_0\right)\ge 0$, user $i$ can always choose $\pi^*\left(\theta,\eta_0\right)=0$ to further increase $V^*\left(\theta,\eta_0\right)$. On the other hand, if $-c+\delta V^*\left(\theta+1,\eta_0\right)<0$, then user $i$ can always choose $\pi^*\left(\theta,\eta_0\right)=L+1$ to further increase $V^*\left(\theta,\eta_0\right)$. Therefore, it can be concluded that user $i$ will never choose the service threshold $h$ in its optimal policy and hence $\underline{k}=\overline{k}$, whose value is denoted as $k$ for short. Finally by substituting $V^*\left(\theta,\eta_0\right)$ into the representation, we have the optimal policy and the corresponding optimal value function for the bad user $i$ as follows:

$$\pi^*\left(\theta,\eta_0\right)=\begin{cases}L+1, & if\ 0\le\theta\le k\\ 0, & if\ k+1\le\theta\le h-1\end{cases},V^*\left(\theta,\eta_0\right)=\begin{cases}0, & if\ 0\le\theta\le k\\ \delta^{h-\theta}V^*\left(h,\eta_0\right)-\dfrac{1-\delta^{h-\theta}}{1-\delta}c, & if\ k+1\le\theta\le h-1\end{cases}.\quad(21)$$

Using the one-shot deviation principle [11], user $i$ chooses to play $a=L+1$ instead of $a=0$ when $\theta\le k$ if and only if the following condition is satisfied:

$$c\ge\delta\left[V^*\left(\theta+1,\eta_0\right)-V^*\left(0,\eta_0\right)\right],\ for\ \theta\le k.\quad(22)$$

Since $V^*\left(\theta,\eta_0\right)$ monotonically increases against $\theta$, we only need to consider $V^*\left(k+1,\eta_0\right)-V^*\left(0,\eta_0\right)$, i.e. $c\ge\delta\left[\delta^{h-k-1}V^*\left(h,\eta_0\right)-\dfrac{1-\delta^{h-k-1}}{1-\delta}c\right]$. With a reputation of 0, user $i$ is enforced to play $\sigma_D$ in its best response as long as $k\ge 0$, which is equivalent to the following inequality

$$c\ge\delta^hV^*\left(h,\eta_0\right)-\frac{\delta-\delta^h}{1-\delta}c.\quad(23)$$

Therefore, user $i$'s incentive to play $\sigma_D$ in its best response depends on the value of $V^*\left(h,\eta_0\right)$. When $p_0\left(0\right)\le\dfrac{\delta b-c}{\delta\left(b-c\right)}$, $\pi_0^*\left(\theta,\eta_0\right)=h$ for $\theta\ge h$ and $V^*\left(h,\eta_0\right)=\dfrac{p_0\left(L\right)\left(b-c\right)}{1-\delta}$; and when $p_0\left(0\right)\ge\dfrac{\delta b-c}{\delta\left(b-c\right)}$, $\pi_0^*\left(\theta,\eta_0\right)=L+1$ for $\theta\ge h$ and $V^*\left(h,\eta_0\right)=\dfrac{p_0\left(L\right)b}{1-\delta p_0\left(0\right)}$.

We first consider the case when $p_0\left(0\right)\le\dfrac{\delta b-c}{\delta\left(b-c\right)}$ and $V^*\left(h,\eta_0\right)=\dfrac{p_0\left(L\right)\left(b-c\right)}{1-\delta}$. Hence, the inequality (22) equals to $c\ge\delta^h\dfrac{p_0\left(L\right)\left(b-c\right)}{1-\delta}-\dfrac{\delta-\delta^h}{1-\delta}c$. Consequently, user $i$ is enforced to play $\sigma_D$ in its best response if and only if $1-\frac{\left(1-\delta^{h}\right)c}{\delta^{h}\left(b-c\right)} \leq p_0\left(0\right) \leq \frac{\delta b-c}{\delta\left(b-c\right)}$. Similarly, when $p_0\left(0\right) \geq \frac{\delta b-c}{\delta\left(b-c\right)}$ and $V^*\left(h,\eta_0\right)=\frac{p_0\left(L\right)b}{1-\delta p_0\left(0\right)}$, we can prove that user $i$ always has the incentive to play $\sigma_D$ in its best response.

Here we assume that when user $i$ expects to receive the same utility by choosing $\sigma_D$ and $\phi$, it will choose either of the two options with probability 0.5. Hence, we sum up the condition for user $i$ being enforced to play $\sigma_D$ in its best response, i.e. $\pi^*\left(0,\eta_0\right)=L+1$, to be $p_0\left(0\right)>1-\frac{\left(1-\delta^{h}\right)c}{\delta^{h}\left(b-c\right)}$.

*2) Enforcing $\phi$ on users of reputation $L$*

In this section, we analyze the condition for users of reputation $L$ to comply with $\phi$ in their best responses.

Similarly, a user $j$ of reputation $L$ confronts an opponent configuration $\eta_L=\left\{m_L\left(0\right),\cdots,m_L\left(L\right)\right\}$, with $m_L\left(0\right)=n\left(0\right)$ and $m_L\left(L\right)=n\left(L\right)-1=N-1-m_L\left(0\right)$. We also use $p_L\left(0\right)=\frac{n\left(0\right)}{N-1}$ and $p_L\left(L\right)=\frac{n\left(L\right)-1}{N-1}$ to represent the fractions of users in user $j$ 's opponent configuration, and let $\left\{\pi^*\left(\theta,\eta_L\right)\right\}_{\theta=0}^{L}$ and $V^*\left(\theta,\eta_L\right)$ denote its optimal policy and value function. Similarly, there exists $k' \in \left\{-1,0,\ldots,h-1\right\}$ and we have that

$$\pi^*\left(\theta,\eta_L\right)=\begin{cases}L+1, & if\ 0\leq\theta\leq k' \\ 0, & if\ k'+1\leq\theta\leq h-1\end{cases}, V^*\left(\theta,\eta_L\right)=\begin{cases}0, & if\ 0\leq\theta\leq k' \\ \delta^{h-\theta}V^*\left(h,\eta\right)-\frac{1-\delta^{h-\theta}}{1-\delta}c, & if\ k'+1\leq\theta\leq h-1\end{cases}. \quad (24)$$

Using the same approach as that in the previous section, we can derive the condition for user $j$ being enforced to comply with $\phi$ in its best response, i.e. $\pi^*\left(L,\eta_L\right)=h$, to be $p_L\left(L\right)>\frac{\left(1-\delta\right)c}{\delta\left(b-c\right)}$.

*3) The sufficient and necessary condition for an absorbing configuration*

Combining the results from the above two sections, we can now discuss the sufficient and necessary condition for a configuration to be absorbing.

We first analyze the configuration with all users having the reputation 0, i.e. $n(0) = N$ and $p_0(0) = 1$. This configuration is absorbing if and only if each user has the incentive to play $\sigma_D$ in its best response. Since $1 > 1 - \frac{(1-\delta^h)c}{\delta^h(b-c)}$, the condition is satisfied and this is always an absorbing configuration.

We then analyze the configuration with all users having the reputation $L$, i.e. $n(L) = N$ and $p_L(L) = 1$. Since $1 > \frac{(1-\delta)c}{\delta(b-c)}$ if and only if $\delta b > c$, this is an absorbing configuration if and only if $\delta b > c$.

Finally, we analyze the general configuration $\mu$ with $n(0) > 0$, $n(L) > 0$, and $n(0) + n(L) = 1$, where we have $p_0(0) \geq 0$, $p_L(L) \geq 0$, and $p_0(0) + p_L(L) = \frac{N-2}{N-1}$. In this case, any configuration that has $\underline{B} < n(L) < \min\{N, \overline{B}\}$ is absorbing, where $\underline{B} = \frac{(1-\delta)c}{\delta(b-c)}(N-1) + 1$ and $\overline{B} = \frac{(1-\delta^h)c}{\delta^h(b-c)}(N-1)$. It should also be noted that when $0 < n(L) \leq \underline{B}$, users of reputation 0 has sufficient incentives to follow $\sigma_D$, while users of reputation $L$ loses the incentive to comply with $\phi$. As a result, the community configuration will converge to $\mu_0$. Similarly, a community configuration with $n(L) \geq \overline{B}$ is not absorbing and will converges to $\mu_N$.

Therefore, there is always at least one community configuration that is an absorbing configuration.

*B. The analysis of absorbing classes*

In this section, we prove that any irreducible absorbing class should contain at least one absorbing configuration. Since an absorbing configuration is also a degenerated absorbing class, this in turn implies that each irreducible absorbing class is an absorbing configuration and there is no irreducible absorbing class that contains more than one configuration.

Let us consider an irreducible absorbing class $M$ that is non-empty and contains at least one configuration is not an absorbing configuration. If the non-absorbing configuration satisfies (11), there are only users of reputation 0 and $L$. Since this configuration is non-absorbing, we should have either users of reputation 0 choose $\sigma_C$ in their best responses or users of reputation $L$ choose $\sigma_D$ in their best responses. In either case, this non-absorbing configuration has a positive probability to transit to an absorbing configuration. If the non-absorbing configuration does not satisfy (11), it is obvious that this configuration has a positive probability to transit into a configuration satisfying (11) since each user has a positive probability $1-\gamma$ to continue play with the same service strategy.

Hence, any non-absorbing configuration has a positive probability to transit into an absorbing configuration. Since an absorbing class should be communicating, it should contain at least one absorbing configuration. Therefore, the statement at the beginning of this section follows.

*C. Conclusion*

Summing up the above analysis, we can conclude that each stochastically stable configuration is an absorbing configuration which satisfies (11). On the other hand, since each stochastically stable configuration is absorbing under the best response dynamics, it should also belong to a stochastically stable equilibrium. Therefore, Theorem 1 follows. ■

## APPENDIX C

## PROOF OF THEOREM 2

Theorem 1 shows that it is sufficient to consider configurations satisfying (11) to in the analysis of stochastically stable configurations. There are three sets of absorbing configurations: (a) $\mu_0 = \{n(0) = N, n(L) = 0\}$ ; (b) $\mu_N = \{n(0) = 0, n(L) = N\}$ ; (c) $\mu = \{n(0), n(L)\}$ with $\underline{B} \le n(L) \le \min\{N-1, \overline{B}\}$ and $n(0) = N - n(L)$.

As shown in Theorem 1, Set (a) is always non-empty, Set (2) is non-empty if and only if $\frac{c}{b} < \delta$. Hence, $\mu_N$ could be a stochastically stable equilibrium only if $\frac{c}{b} < \delta$ is satisfied.

When $\frac{c}{b} < \delta$, we analyse Set (3).

(1) If $\overline{B} > N - 1$, $\mu_N$ is not the unique stochastically stable equilibrium due to the fact that one occurrence of operation error is needed for a user to transit from the reputation $L$ to the reputation 0 and $h$ occurrence of operation error is needed for a user to transit from the reputation 0 to the reputation $L$ . Since $h \ge 1$, there is at least one other absorbing configuration which has a basin of attraction no smaller than $\mu_N$.

(2) If $\underline{B} \ge 1$ and $\overline{B} \le N - 1$, it is easy to show that $\mu_0$ has a basin of attraction that is larger than any absorbing configurations other than $\mu_N$. Hence, we only have to compare the basins of attraction of $\mu_0$ and $\mu_N$. The number of occurrence of operation errors for $\mu_0$ to move into $\mu_N$ is $\overline{B}h$, and the number of occurrence of operation errors for $\mu_N$ to move into $\mu_0$ is $N - \underline{B}$. Hence, the condition for $\mu_N$ to be the unique stochastically stable equilibrium is that $N - \underline{B} > \overline{B}h$.

To sum up, the condition for $\mu_N$ to be the unique stochastically stable equilibrium is that $N - \underline{B} > \overline{B}h$ and $\overline{B} < N - 1$, which can be translated into the following inequality as

$$\delta^h b - \delta^{h-1} c \ge \left(1 - \delta^h\right) ch . \tag{25}$$

Since the LHS of (25) monotonically decreases against $h$, while the RHS of (25) monotonically increases with $h$, we can conclude that there exists an $H$ such that (25) is satisfied when $h < H$. ■

## APPENDIX D

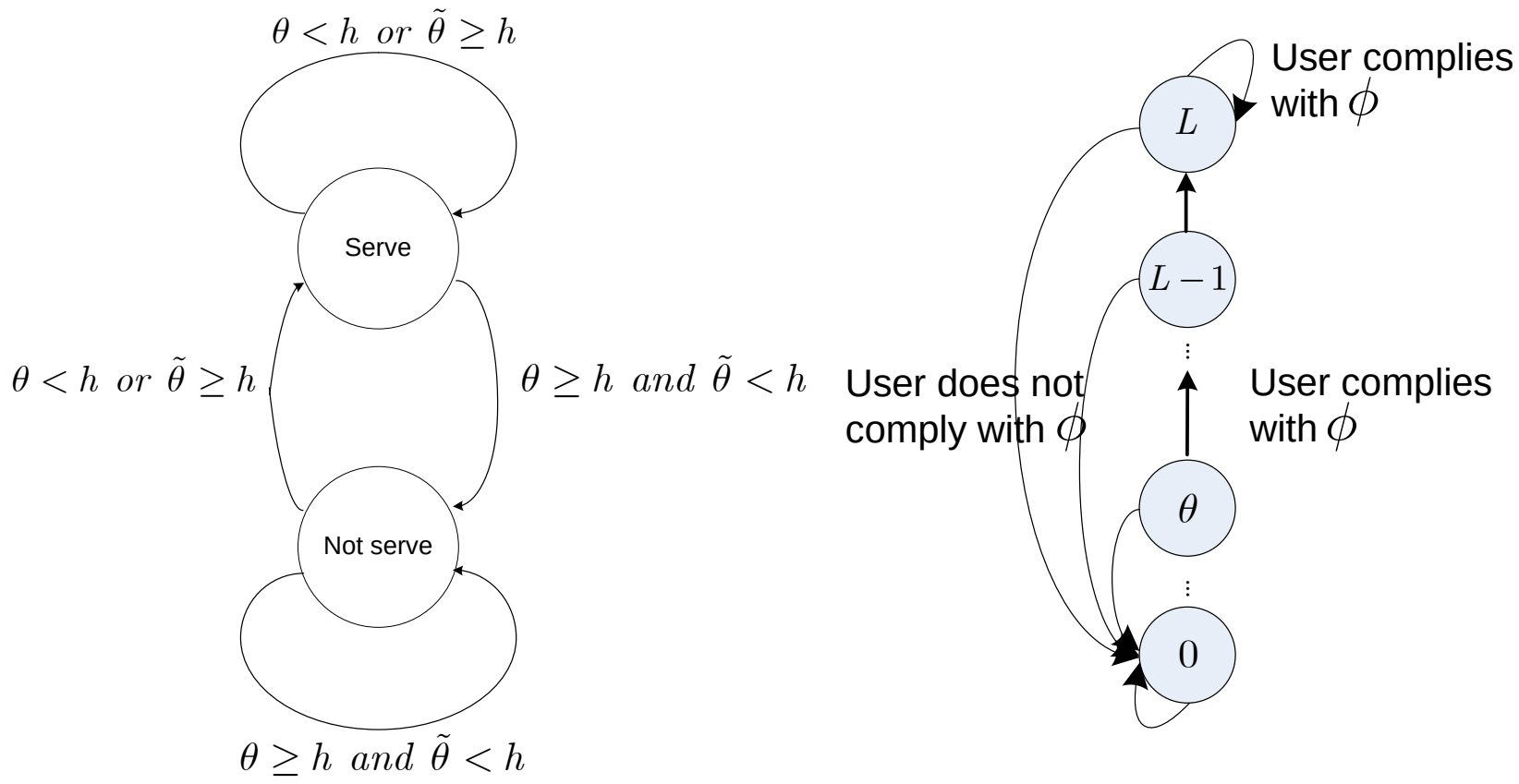

Figure 1. The schematic representation of the social rule and the reputation scheme

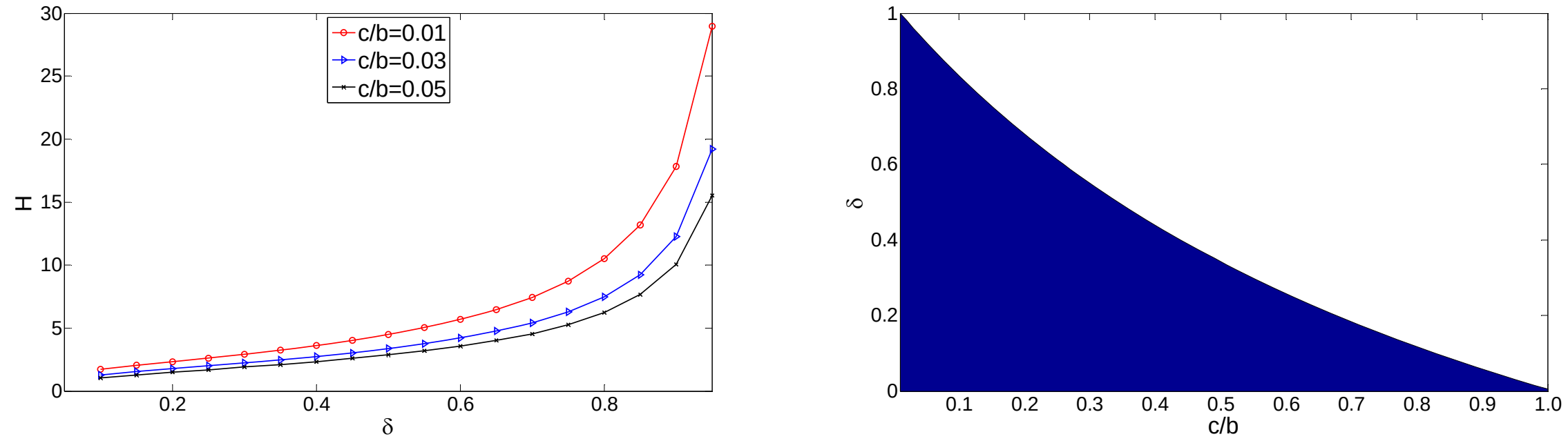

(a) (b)

Figure 2. (a) $H$ against $\delta$ and $c/b$; (b) The feasible region of $\delta$ and $c/b$;

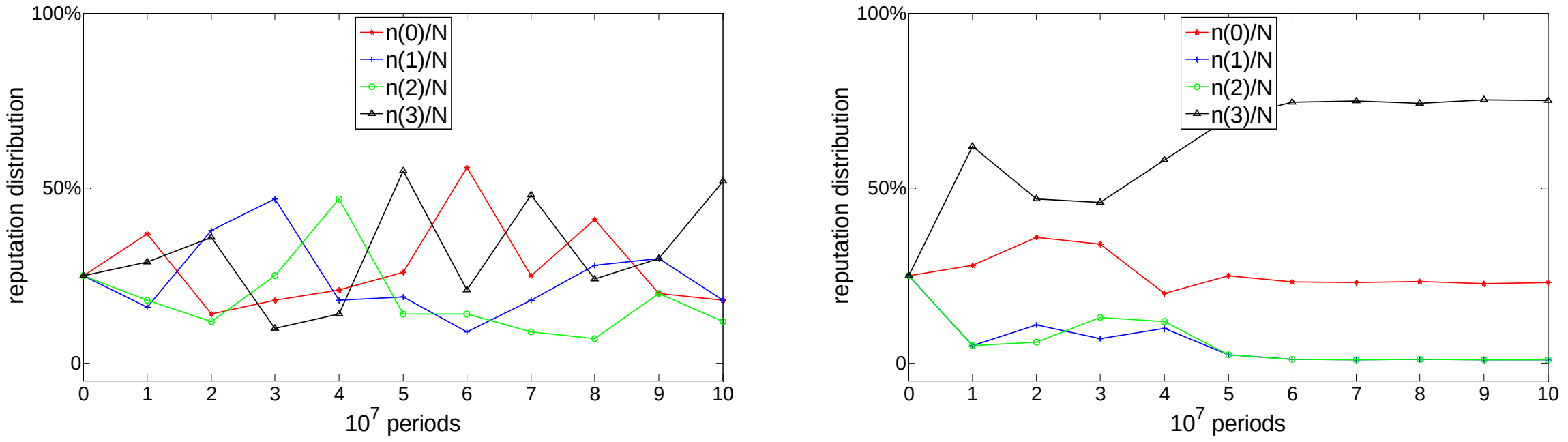

(a) $\varepsilon = 0.2$ and $b = 3$ (b) $\varepsilon = 0.05$ and $b = 3$

Figure 3. The evolution of the community in $10^8$ periods. ( $c = 1, \delta = 0.5$ )

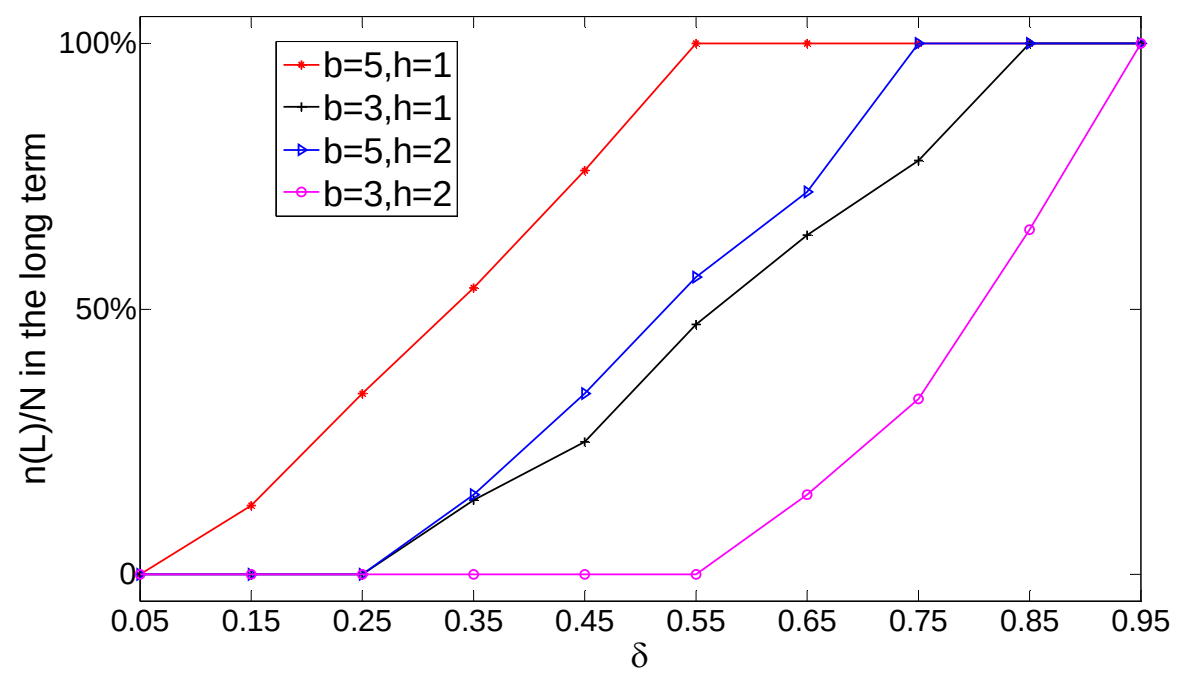

Figure 4. The fraction $n(L)$ of users of reputation $L$ after $10^8$ periods. ( $c=1$ )

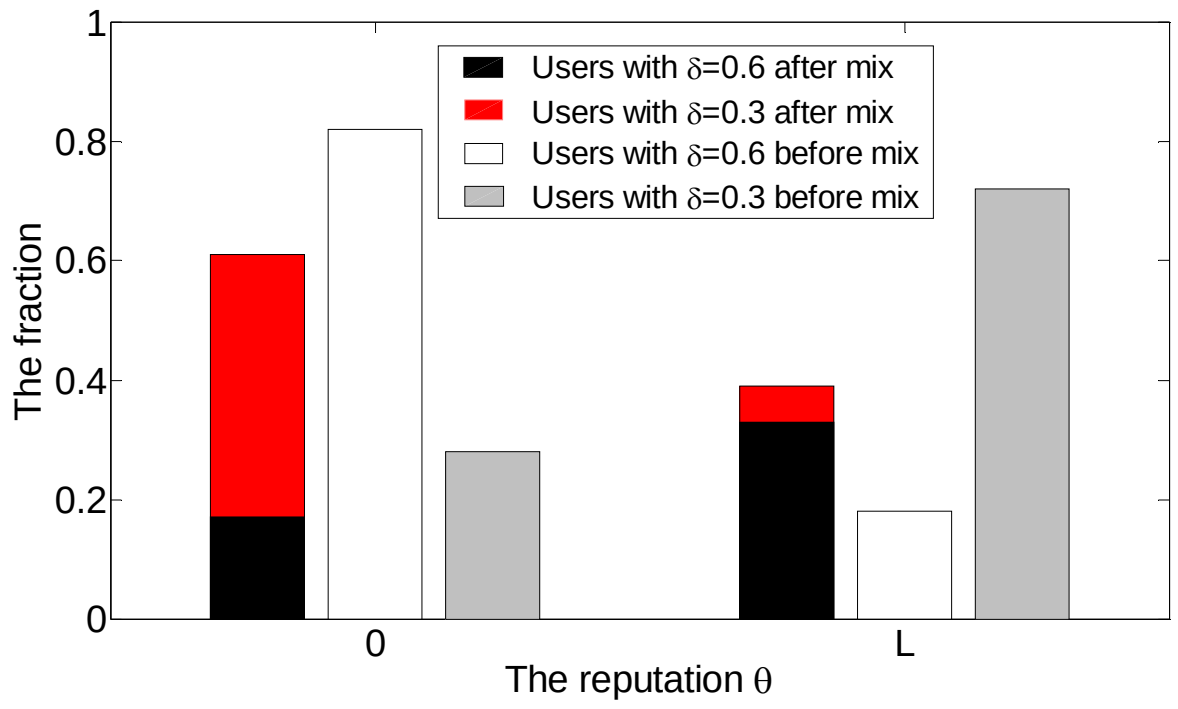

Figure 5. The comparison of reputation distributions in the long-term
( $L=3, h=2, \varepsilon=0.05$, $c=1$, $b=2$, $N=500$ )

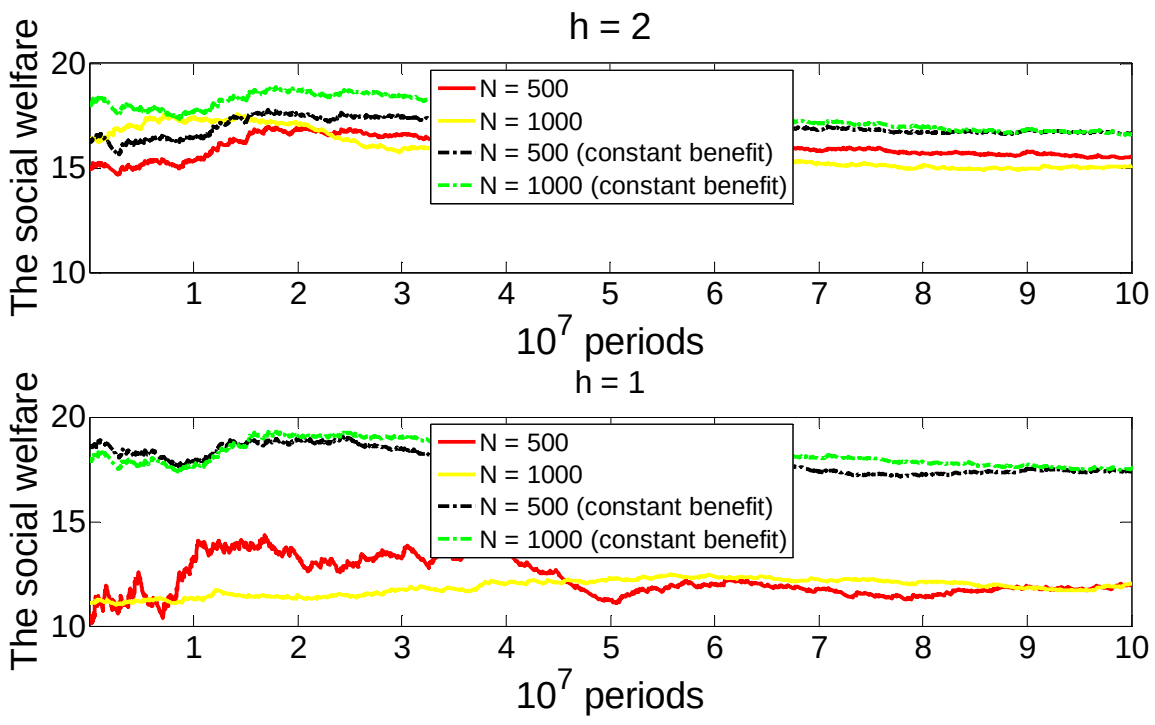

Figure 6. The evolution of the social welfare in the long term when the stage game benefit varies over time
( $L=3, \delta=0.5, \varepsilon=0.05$, $c=1$ )

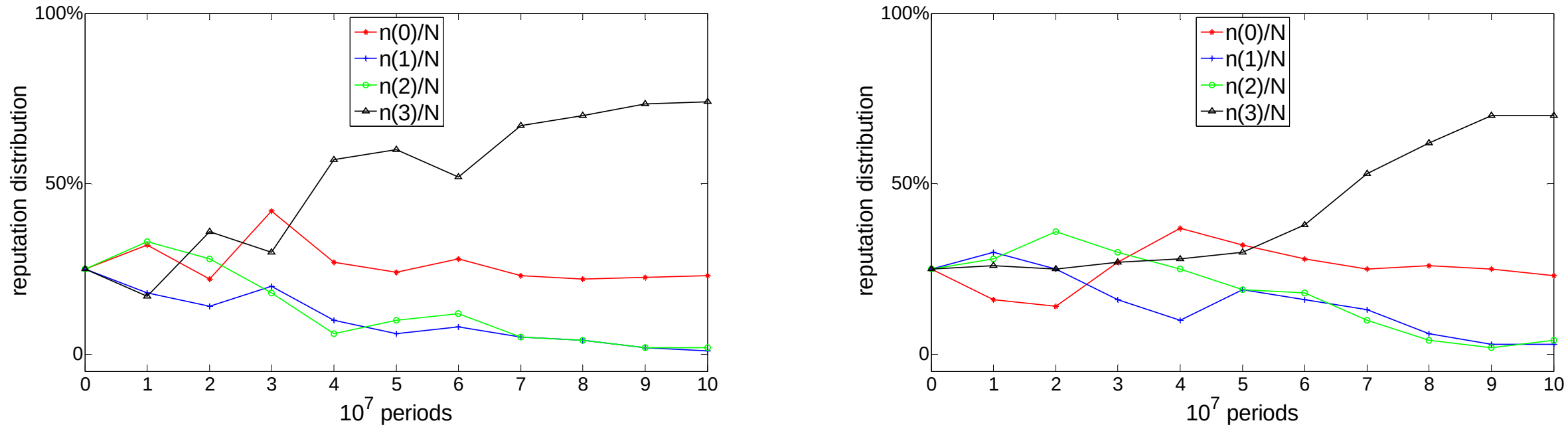

(a) $N=500$ (b) $N=1000$

Figure 7. The evolution of the community in $10^8$ periods.
( $\varepsilon=0.05, b=3, c=1, \delta=0.5$ )